\documentclass{bmvc2k}
\usepackage{multirow}
\usepackage{graphicx}
\usepackage{booktabs}
\usepackage{epsfig}
\usepackage{amsmath}
\usepackage{amssymb}
\usepackage{booktabs}
\usepackage{enumitem}
\usepackage{bbm}
\usepackage{tikz}
\usepackage{makecell}
\usetikzlibrary{arrows.meta,positioning,calc,fit,backgrounds}
\usepackage{algorithm}
\usepackage{algpseudocode}

\algnewcommand\algorithmicinput{\textbf{Input:}}
\algnewcommand\Input{\item[\algorithmicinput]}
\algnewcommand\algorithmicoutput{\textbf{Output:}}
\algnewcommand\Output{\item[\algorithmicoutput]}

\title{
ForestMamba: Sparse Mamba with Geometry-guided Queries for 3D Forest Point Cloud Segmentation
}

\addauthor{Trung Thanh Nguyen}{nguyent@cs.is.i.nagoya-u.ac.jp}{1,2}
\addauthor{Tuan-Anh Vu}{tuananh.vu@ucla.edu}{3}
\addauthor{Duc Viet Le}{v.d.le@utwente.nl}{4}
\addauthor{Yasutomo Kawanishi}{kwns@fc.ritsumei.ac.jp}{5,2,1}
\addauthor{Takahiro Komamizu}{taka-coma@acm.org}{1}
\addauthor{Ichiro Ide}{ide@i.nagoya-u.ac.jp}{1}
\addauthor{Teja Kattenborn}{teja.kattenborn@geosense.uni-freiburg.de}{6}

\addinstitution{
 Nagoya University\\
 Nagoya, Aichi, Japan
}
\addinstitution{
 RIKEN\\
 Seika, Kyoto, Japan
}

\addinstitution{
 University of California, Los Angeles\\
 CA, USA
}

\addinstitution{
 University of Twente\\
 Enschede, Overijssel, the Netherlands
}

\addinstitution{
 Ritsumeikan University \\
 Ibaraki, Osaka, Japan
}

\addinstitution{
 University of Freiburg\\
 Freiburg im Breisgau, BW, Germany
}

\runninghead{Trung Thanh Nguyen et al.}{ForestMamba}

\begin{document}
\begingroup
\setlength{\leftskip}{0.7cm}
\maketitle
\endgroup
\maketitle

\begin{abstract}
Semantic and instance segmentation of terrestrial and drone LiDAR point clouds is emerging as a transformative approach for converting the complex 3D structure of forests into actionable information for forest monitoring and biodiversity assessment.
However, forest LiDAR scenes remain highly challenging due to their large data volumes, irregular sampling density, overlapping and complex canopy structure, and geographic variability.
Existing methods based on sparse convolutions or Transformers achieve promising results, but suffer from two key limitations: Quadratic complexity of attention scales poorly to large forest scenes, and generic context modeling does not exploit forest structural priors, limiting tree separation in complex regions.
To address these challenges, we propose \textbf{ForestMamba}, a structure-aware method that incorporates forest-specific priors into feature encoding, query generation, and query refinement, while replacing quadratic attention with linear-time state-space modeling.
First, we introduce a sparse encoder with vertical-priority slab serialization that organizes sparse voxels into vertically coherent sequences for efficient long-range context modeling.
Second, we propose a geometry-guided query initialization strategy based on an on-the-fly multi-scale Canopy Height Model (CHM), where canopy maxima provide ecologically meaningful query seeds, supplemented by Farthest Point Sampling (FPS) to cover understory trees.
Third, we design a Mamba-based query decoder that combines local $k$NN voxel aggregation with a spatial dual-path Mamba for query refinement with linear computational complexity.
Extensive experiments across seven geographically distributed forest regions demonstrate that the proposed ForestMamba outperforms existing baselines overall in both instance and semantic segmentation, while achieving approximately $3\times$ faster inference and $2.3\times$ lower GPU memory consumption.
These results suggest that combining structural priors with linear-time sequence modeling is promising for large-scale forest scene understanding.
\end{abstract}

\section{Introduction}
\label{sec:intro}

Forests cover approximately 31\% of the Earth's land surface and play a central role in carbon cycling, biodiversity conservation, and climate regulation~\cite{white2016remote,duncanson2022aboveground}.
Forests are inherently complex 3D ecosystems, and the combination of terrestrial and drone-based LiDAR with semantic and instance segmentation is unlocking new possibilities for effective forest monitoring and biodiversity assessment~\cite{calders2020terrestrial,brede2022nondestructive, kattenborn2021review,nguyen2026selectanytree}.
Ecological applications depend on accurately delineating individual trees for analyzing biomass, crown architecture, and competition, while simultaneously distinguishing semantic components such as foliage, woody material, and terrain to characterize forest productivity and health. Joint semantic and instance segmentation addresses these complementary objectives within a unified framework.
Compared with indoor 3D scenes, point clouds of forest environments are substantially more challenging.
They are extremely large, often containing hundreds of millions of points per hectare at high acquisition densities~\cite{xiang2024automated,calders2020terrestrial}, exhibit irregular sampling density due to occlusion and sensor geometry~\cite{wielgosz2024segmentanytree}, and contain complex, interlocking canopy structures~\cite{xiang2025forestformer3d}. 
In addition, severe class imbalance among ground, woody components, and foliage further complicates segmentation~\cite{xiang2024automated}. 
These factors make accurate segmentation in forests considerably more difficult than in conventional 3D vision benchmarks.

Recent advances in deep learning have significantly improved 3D point cloud segmentation, with sparse convolutional and point-based architectures enabling scalable processing of large scenes~\cite{choy2019minkowski,graham2018spconv}.
In forest environments, early methods such as ForAINet~\cite{xiang2024automated} employ sparse convolutional U-Net backbones~\cite{ronneberger2015u} combined with post-hoc clustering to group points into individual trees.
Subsequent methods largely follow a similar paradigm, including SegmentAnyTree~\cite{wielgosz2024segmentanytree}, which extends this line of work to class-agnostic tree segmentation, but still relies on heuristic grouping strategies such as mean-shift or Density-Based Spatial Clustering of Applications with Noise (DBSCAN)~\cite{10.5555/3001460.3001507}, as in TreeLearn~\cite{henrich2024treelearn}.
While computationally efficient, these methods are fundamentally limited by their restricted receptive fields and dependence on hand-crafted clustering, making them brittle in dense, overlapping canopy structures where tree instances are highly entangled.

More recently, Transformer-based methods~\cite{kolodiazhnyi2024oneformer3d,cheng2022masked} have been introduced to capture long-range dependencies in 3D point cloud segmentation.
Methods such as ForestFormer3D~\cite{xiang2025forestformer3d} leverage global attention and mask-based query decoders to achieve strong instance segmentation performance.
However, these methods face two key limitations:
First, the quadratic complexity of self- and cross-attention with respect to the number of voxels and queries leads to slow inference and training, high memory consumption, and poor scalability to large forest scenes, forcing reliance on extensive spatial partitioning and heavy post-processing, which further slows end-to-end deployment.
Second, generic attention mechanisms do not explicitly encode the structural priors of forest environments, and existing query initialization strategies are agnostic to forest geometry, often producing redundant queries in dense regions while missing weak or partially occluded trees.

To address these limitations, we propose \textbf{ForestMamba}, a structure-aware method for 3D forest point cloud segmentation that injects domain-specific priors into feature encoding and query generation, while replacing quadratic attention with Mamba~\cite{gu2024mamba}, a State Space Model (SSM) that processes sequences via linear recurrence rather than pairwise attention.
It exploits the inherent vertical organization of forest scenes, where trees exhibit a consistent ground-to-canopy hierarchy.
Building on this insight, we introduce a structure-aware sparse encoder with a vertical-priority slab serialization that organizes sparse voxels into vertically coherent sequences, enabling efficient long-range context modeling while preserving the multi-scale representation of a sparse U-Net backbone~\cite{ronneberger2015u}.
We further propose a geometry-guided query initialization strategy based on an on-the-fly multi-scale Canopy Height Model (CHM), in which instance queries are seeded from local canopy maxima and enriched via cylinder-based feature pooling, with Farthest Point Sampling (FPS)~\cite{xiang2025forestformer3d} over auxiliary discriminative embeddings supplementing the queries to cover understory trees.
The resulting queries are refined by a Mamba-based query decoder that combines local $k$NN voxel aggregation with a spatial dual-path Mamba over query sequences, achieving linear computational complexity.
The main contributions of this work are summarized as follows:
\begin{itemize}
    \item We propose ForestMamba, a structure-aware method for joint semantic and instance segmentation of 3D forest point clouds that explicitly incorporates the vertical structure of forest scenes into both feature encoding and query generation.
    \item We introduce three key components: (1) Sparse encoder with vertical-priority slab serialization, (2) Geometry-guided query initialization based on multi-scale CHM peak detection supplemented by FPS, and (3) Mamba-based query decoder with local $k$NN aggregation and dual-path spatial scanning.
    \item We conduct comprehensive evaluations across seven geographically distributed forest regions, demonstrating consistent improvements over State-Of-The-Art (SOTA) methods in segmentation accuracy, with $3\times$ faster inference and $2.3\times$ lower GPU memory than Transformer-based methods.
\end{itemize}

\section{Related work}
\label{sec:related_work}

\noindent \textbf{3D point cloud segmentation.}
Forest point clouds can be acquired from airborne, terrestrial, or drone-based LiDAR. 
Accurate semantic and instance segmentation nevertheless relies on dense close-range scans that capture stems, branches, and foliage across multiple forest layers~\cite{kattenborn2021review, feigl2025close}. 
Recent deep learning methods have progressed from point-based networks to sparse convolutional and attention-based architectures.
PointNet~\cite{qi2017pointnet} and PointNet++~\cite{qi2017pointnetpp} pioneered direct point-level learning, while Kernel Point Convolution (KPConv)~\cite{thomas2019kpconv} introduced continuous kernels adapted to local geometry. 
Sparse voxel convolution frameworks such as Sparse Convolutional (SpConv)~\cite{graham2018spconv} and MinkowskiNet~\cite{choy2019minkowski} enabled efficient large-scale 3D processing and have become a standard backbone for scene understanding, with additional robustness improvements explored through test-time augmentation~\cite{vu2024testtime}. 
More recently, Transformer-based models, including the Point Transformer family~\cite{zhao2021point,wu2022ptv2,wu2024ptv3}, have improved long-range context modeling through attention over serialized or local neighborhoods. 
In parallel, State Space Models (SSMs) have emerged as an efficient alternative for long-sequence modeling. 
Following advances such as Mamba~\cite{gu2024mamba} and VMamba~\cite{liu2024vmamba}, several works have begun extending SSMs to 3D point clouds~\cite{liang2024pointmamba,han2024mamba3d,wang20253dumamba,zhang2024pointmamba2}. 
However, existing 3D SSMs focus on general-purpose tasks without adapting to forest-specific structural priors. 
The proposed ForestMamba addresses this by introducing a structure-aware encoder and Mamba-based decoder tailored to the vertical organization of trees.

\vspace{5pt}
\noindent \textbf{Forest point cloud segmentation.}
Individual tree segmentation from LiDAR data has long been studied in remote sensing, where classical methods often rely on Canopy Height Models (CHMs), local maxima detection, and watershed-style region growing~\cite{naveed2019individual,popescu2002chm}. 
Although effective for dominant and well-separated trees, these approaches often struggle in dense forests with multi-layered and overlapping crowns. 
In these situations, recent deep learning methods have substantially improved performance. 
TreeLearn~\cite{henrich2024treelearn} and SegmentAnyTree~\cite{wielgosz2024segmentanytree}, among others~\cite{henrich2024generaltree,rizaldy2025labelefficient3dforestmapping,xiu20253dpsnet}, adapt modern point cloud segmentation frameworks to forest data, but many still depend on heuristic grouping, non-end-to-end pipelines, or manual prompting. 
ForestFormer3D~\cite{xiang2025forestformer3d} further advances the field by introducing a query-based forest segmentation framework with strong empirical performance. 
Nevertheless, current methods still face challenges in modeling long-range forest structure and robustly separating highly entangled tree instances. 
The proposed ForestMamba addresses these issues by integrating forest-specific priors into feature encoding and query generation.

\vspace{5pt}
\noindent \textbf{Query-based instance segmentation.}
Query-based set prediction, popularized by DEtection TRansformer (DETR)~\cite{carion2020detr}, has become a powerful paradigm for instance segmentation. Mask2Former~\cite{cheng2022masked} extends this idea to universal segmentation through masked attention, and OneFormer3D~\cite{kolodiazhnyi2024oneformer3d} brings unified query-based segmentation to 3D scenes. 
In forest point clouds, ForestFormer3D~\cite{xiang2025forestformer3d} adapts this paradigm with a Transformer decoder and one-to-many matching to better handle large scenes and crop boundaries. 
However, existing query initialization strategies are largely generic, relying on learned embeddings or feature-space sampling such as Farthest Point Sampling (FPS)~\cite{xiang2025forestformer3d}, which do not explicitly reflect forest geometry. 
ForestMamba addresses this with a geometry-guided strategy based on an on-the-fly CHM, where canopy maxima provide ecologically meaningful seeds enriched by cylinder-based feature pooling, supplemented by FPS to improve instance coverage.

\section{Proposed method: ForestMamba}
\label{sec:proposed_method}
\noindent \textbf{Problem formulation.}
Given a forest point cloud $\mathcal{P}=\{\mathbf{p}_i \in \mathbb{R}^{3}\}_{i=1}^{N}$, 
where $N$ is the number of points, the goal is to jointly predict a semantic label 
and an instance assignment for each point.
Let $y_i \in \{1,\dots,C\}$ denote the semantic class of point $\mathbf{p}_i$, 
where $C$ is the number of semantic classes, and let $z_i \in \{0,1,\dots,M\}$ 
denote its instance label, where $z_i=0$ denotes non-instance background (e.g., ground) 
and $M$ is the number of tree instances in the scene.
We aim to learn a function that simultaneously performs semantic segmentation and instance-level tree delineation as:
\begin{equation}
f: \mathcal{P} \mapsto \left(\{\hat{y}_i\}_{i=1}^{N}, \{\hat{z}_i\}_{i=1}^{N}\right).
\end{equation}
This task is particularly challenging in forest scenes for three reasons.
First, forest point clouds are large-scale and sparsely distributed, making dense full-scene processing inefficient.
Second, tree instances are often highly entangled in overlapping canopies, complicating instance separation.
Third, generic query initialization strategies (e.g., learned embeddings or Farthest Point Sampling (FPS)~\cite{xiang2025forestformer3d}) do not explicitly capture the geometric structure of trees.
To address these challenges, we propose \textbf{ForestMamba}, a structure-aware method that incorporates forest-specific priors into feature encoding and instance 
query generation.

\begin{figure}[t]
  \centering
  \includegraphics[width=\linewidth]{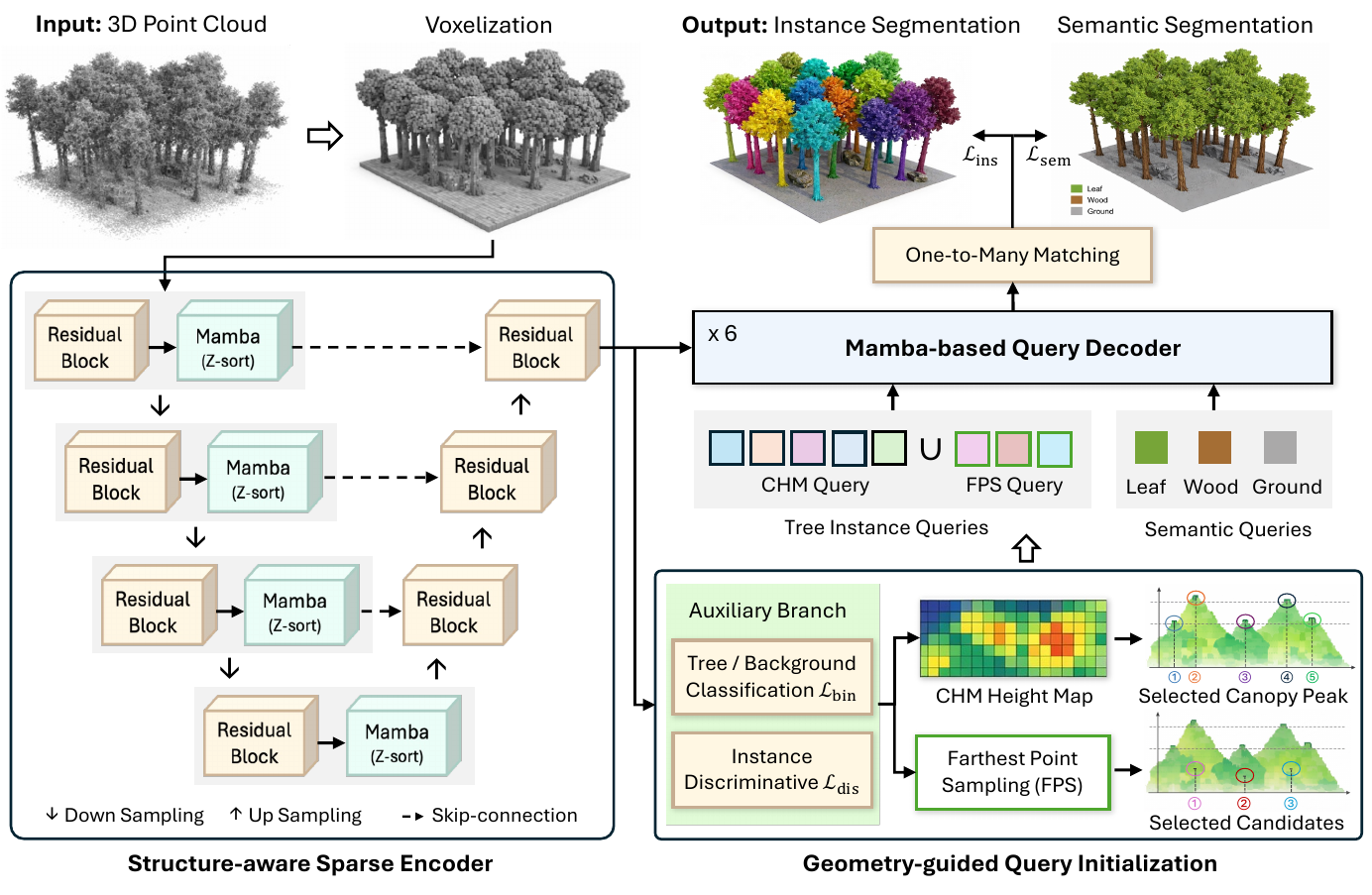}
    \vspace{-10pt}
    \caption{\textbf{Overview of the proposed ForestMamba.}
    The input point cloud is voxelized and processed by a structure-aware sparse encoder with vertical-priority serialization.
    A geometry-guided module initializes instance queries using Canopy Height Model (CHM)-based canopy peaks and Farthest Point Sampling (FPS).
    Resulting queries are refined by a Mamba-based query decoder to jointly predict semantic labels and tree instances.
    }
  \label{fig:architecture}
\end{figure}

\vspace{5pt}
\noindent \textbf{Method overview.}
As illustrated in Fig.~\ref{fig:architecture}, ForestMamba consists of three main components: (1) Structure-aware sparse encoder, (2) Geometry-guided query initialization module, and (3) Mamba-based query decoder.
The input point cloud is first partitioned into cylindrical regions and voxelized with voxel size $v$, yielding $N_v$ occupied voxels (where $N_v \leq N$), which are then processed by a U-Net-style sparse encoder to produce voxel features.
An auxiliary branch predicts tree or background logits, as well as instance-discriminative embeddings.
Based on predicted tree voxels, we construct an on-the-fly Canopy Height Model (CHM), detect canopy peaks, and pool local features to form geometry-guided instance queries, supplemented by FPS to cover understory or occluded trees.
Finally, the queries are refined by a Mamba-based decoder to jointly predict instance masks and semantic labels.

\subsection{Structure-aware sparse encoder}
\label{sec:sparse_mamba}
\noindent \textbf{Motivation.}
Trees exhibit a strong vertical organization from the ground to the trunk, branches, and canopy.
Generic spatial serializations, such as space-filling curves or unordered local neighborhoods, do not explicitly encode this ecological prior.
We therefore design a vertical-priority serialization for sparse voxels that models long-range dependencies in a ground-to-canopy-consistent order.

\vspace{5pt}
\noindent \textbf{U-Net-style architecture.}
The encoder follows a U-Net-style architecture~\cite{ronneberger2015u} with $L_e$ downsampling stages, each consisting of a sparse convolutional residual block followed by a Mamba block with slab-ordered serialization.
The decoder mirrors this structure with $L_e$ upsampling stages, where sparse convolutional residual blocks progressively recover spatial resolution.
Skip connections transfer encoder features directly to the corresponding decoder stages, preserving fine-grained geometric details across scales.
The final encoder output is the feature tensor defined as:
\begin{equation}
\mathbf{F}=\{\mathbf{f}_n\}_{n=1}^{N_v}, \qquad \mathbf{f}_n \in \mathbb{R}^{D},
\end{equation}
where $D$ is the output channel dimension. 

\vspace{5pt}
\noindent \textbf{Slab-ordered serialization.}
For each occupied voxel $n$ with integer coordinates $\mathbf{c}_n=(c_n^x, c_n^y, c_n^z)$, we assign it to a vertical slab indexed by:
\begin{equation}
s_n = \left\lfloor \frac{c_n^z}{\tau} \right\rfloor,
\end{equation}
where $\tau \in \mathbb{Z}_{+}$ denotes the slab thickness in voxel units.
We then sort voxels lexicographically by the key $(s_n,\, c_n^y,\, c_n^x)$, yielding a permutation $\pi:\{1,\dots,N_v\}\!\to\!\{1,\dots,N_v\}$ and a serialized feature sequence as:
\begin{equation}
    \tilde{\mathbf{X}}^{(\ell)} = \left[\mathbf{x}_{\pi(1)}^{(\ell)}, 
    \mathbf{x}_{\pi(2)}^{(\ell)}, \dots, \mathbf{x}_{\pi(N_v)}^{(\ell)}\right],
\end{equation}
where $\mathbf{x}_n^{(\ell)} \in \mathbb{R}^{d_\ell}$ denotes the voxel feature at 
encoder stage $\ell$.
This serialization preserves local horizontal coherence within each slab while 
prioritizing the vertical structure of trees.

\vspace{5pt}
\noindent \textbf{Mamba block.}
Given a serialized feature sequence $\tilde{\mathbf{X}}^{(\ell)}=[\tilde{\mathbf{x}}_1,
\dots,\tilde{\mathbf{x}}_{N_v}]$, we apply a Mamba block~\cite{gu2024mamba} to model 
long-range dependencies with linear complexity.
Layer normalization and a residual connection are applied as:
\begin{equation}
    {\tilde{\mathbf{X}}^{\prime(\ell)} = \mathrm{Mamba}\!\left(\mathrm{LN}\big(\tilde{\mathbf{X}}^{(\ell)}\big)\right)
    + \tilde{\mathbf{X}}^{(\ell)}.}
\end{equation}
After processing, features are restored to voxel order via $\pi^{-1}$.
This design combines the local geometric inductive bias of sparse convolutions with efficient long-range modeling of vertical tree structure.

\subsection{Geometry-guided query initialization}
\label{sec:query_init}
\noindent\textbf{Motivation.}
Query-based instance segmentation depends critically on the quality of initial queries.
Generic strategies, such as learned embeddings or FPS in the feature space, are agnostic to forest geometry and often produce redundant queries in dense forest scenes.
To obtain meaningful instance seeds, we generate queries from a multi-scale canopy structure, supplemented by FPS to improve instance coverage.

\vspace{5pt}
\noindent{\textbf{Auxiliary branch.}
Given encoded voxel features $\mathbf{F}=\{\mathbf{f}_n\}_{n=1}^{N_v}$, an auxiliary branch processes the voxel features through two parallel Multi-Layer Perceptron (MLP) heads.
The first head predicts binary tree or background logits for each voxel, yielding the set of predicted tree voxels $\mathcal{T} \subseteq \{1,\dots,N_v\}$.
The second head learns instance-discriminative embeddings $\{\mathbf{e}_n \in \mathbb{R}^{D_e}\}_{n=1}^{N_v}$, where $D_e$ is the embedding dimension, such that voxels belonging to the same tree instance are close in the embedding space while those from different instances are pushed apart, supervised by a discriminative loss~\cite{de2017semantic}.
These two outputs, $\mathcal{T}$ and $\{\mathbf{e}_n\}$, are used for CHM construction and FPS supplementation, respectively.}

\vspace{5pt}
\noindent\textbf{Multi-scale CHM peak detection.}
Let $\mathbf{u}_n=(c_n^x, c_n^y) \in \mathbb{Z}^2$ and $h_n = c_n^z  v$ denote the horizontal coordinate and metric height of voxel $n$, respectively.
For each resolution $r \in \mathcal{R}$, we construct a 2D CHM $H^{(r)} \in \mathbb{R}^{W_r \times H_r}$, where each grid cell $(a,b)$ stores the maximum height of tree voxels falling into that cell as:
\begin{equation}
H^{(r)}(a,b) = \max\!\Big\{\, h_n \;\Big|\; n \in \mathcal{T},\; \big\lfloor {v\,}\mathbf{u}_n / r \big\rfloor = (a,b) \Big\},
\end{equation}
with $H^{(r)}(a,b)=-\infty$ for empty cells, where $v\,\mathbf{u}_n$ converts the voxel index to metric coordinates so that binning is consistent with the CHM resolution $r$ given in meters.
Canopy peaks are detected using an allometric suppression window of radius (in cells) as:
\begin{equation}
w_r(h) = \big\lceil \alpha\, h^{\beta} / r \big\rceil,
\end{equation}
where $\alpha, \beta > 0$ are hyper-parameters empirically tuned on the training split, motivated by the ecological observation that crown radius scales as a power law of tree height~\cite{jucker2017allometric,pretzsch2009allometry}.
A cell $(a,b)$ is declared a local maximum if:
\begin{equation}
    H^{(r)}(a,b) \geq H^{(r)}(a',b'),
    \quad \forall\,(a',b') \in \mathcal{N}_{w_r(H^{(r)}(a,b))}(a,b),
\end{equation}
where $\mathcal{N}_\rho(a,b)=\{(a',b'): \|(a',b')-(a,b)\|_\infty \leq \rho\}$.
For each detected local maximum at resolution $r$, we record its 2D world coordinate $\mathbf{m}_k \in \mathbb{R}^2$ and CHM height $H_k = H^{(r)}(a,b)$.
Candidate peaks from all scales are pooled and deduplicated using greedy spatial non-maximum suppression~\cite{neubeck2006efficient}.
Sorting in descending order of $H_k$, a peak $\mathbf{m}_k$ is retained if:
\begin{equation}
    \|\mathbf{m}_k - \mathbf{m}_j\|_2 \geq d_{\min},
    \quad \forall\, \mathbf{m}_j \in \mathcal{P}_{\text{kept}}\;\text{with}\; H_j > H_k,
\end{equation}
where $\mathcal{P}_{\text{kept}}$ denotes the set of already retained peaks.
Let $\{\mathbf{m}_k\}_{k=1}^{K_{\mathrm{CHM}}}$ denote the final set of retained peaks, where $K_{\mathrm{CHM}} \leq K$ is capped by the query budget $K$. 
If more than $K$ peaks are detected, we retain the top-$K$ ranked by CHM height.

\vspace{5pt}
\noindent\textbf{Cylinder feature pooling.}
For each detected peak $\mathbf{m}_k$, we construct a query by aggregating encoder features from nearby tree voxels within a cylinder of radius $r_p$ as:
\begin{equation}
\Omega_k = \left\{ n \in \mathcal{T} \;\middle|\; \|v\,\mathbf{u}_n - \mathbf{m}_k\|_2 \leq r_p \right\},
\end{equation}
\begin{equation}
\mathbf{q}_k^{\mathrm{CHM}} = \frac{1}{|\Omega_k|} \sum_{n \in \Omega_k} \mathbf{f}_n \;\in\; \mathbb{R}^{D},
\end{equation}
where $v$ converts voxel indices to metric coordinates.
This operation captures local crown context and is more robust than selecting a single voxel feature.

\vspace{5pt}
\noindent\textbf{FPS supplementation.}
If $K_{\mathrm{CHM}} < K$, we generate $K - K_{\mathrm{CHM}}$ additional queries via FPS over tree-voxel discriminative embeddings $\mathcal{E}_{\mathrm{tree}}=\{\mathbf{e}_n \mid n \in \mathcal{T}\}$ produced by the auxiliary branch.
Starting from $\mathcal{S}_0=\emptyset$, FPS iteratively selects:
\begin{equation}
i_t = \arg\max_{j \in \mathcal{T} \setminus \mathcal{S}_{t-1}}
\; \min_{k \in \mathcal{S}_{t-1}} \|\mathbf{e}_j - \mathbf{e}_k\|_2,
\quad \mathcal{S}_t = \mathcal{S}_{t-1} \cup \{i_t\},
\end{equation}
with the convention $\min_{k\in\emptyset}\|\cdot\|_2=+\infty$.
Each selected index $i_t$ yields a query feature $\mathbf{q}_t^{\mathrm{FPS}} = \mathbf{f}_{i_t} \in \mathbb{R}^{D}$.
The final query set is obtained as:
\begin{equation}
\mathcal{Q} =
\big\{\mathbf{q}_1^{\mathrm{CHM}},\dots,\mathbf{q}_{K_{\mathrm{CHM}}}^{\mathrm{CHM}}\big\}
\cup
\big\{\mathbf{q}_1^{\mathrm{FPS}},\dots,\mathbf{q}_{K-K_{\mathrm{CHM}}}^{\mathrm{FPS}}\big\},\quad |\mathcal{Q}|=K.
\end{equation}
CHM-guided queries provide ecologically meaningful treetop seeds, while FPS queries improve coverage of understory or occluded trees that may not yield reliable canopy peaks.

\subsection{Mamba-based query decoder}
\begin{figure}[t]
  \centering
  \includegraphics[width=0.90\linewidth]{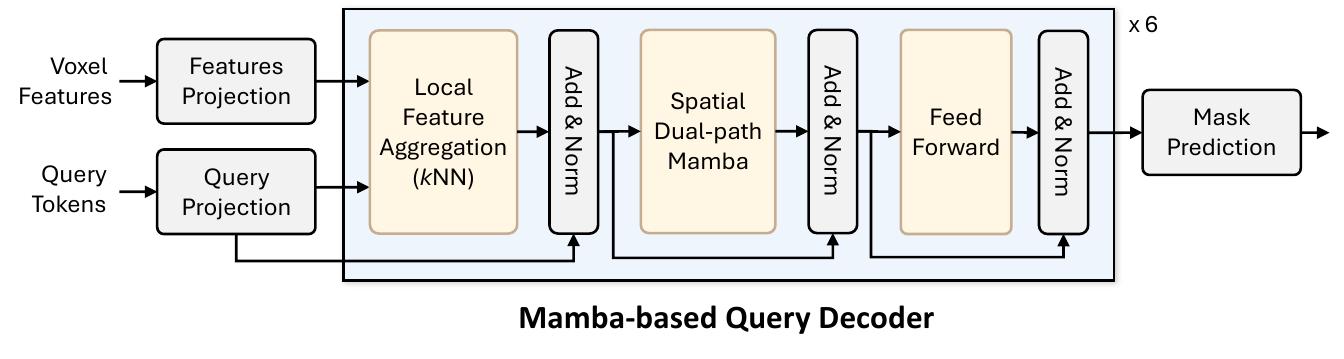}
    \caption{\textbf{Overview of the Mamba-based Query Decoder.}
    Query tokens are iteratively refined through local $k$NN aggregation, spatial Mamba modeling, and feed-forward layers, followed by mask prediction.
    }
  \label{fig:mamba_decoder}
\end{figure}

\noindent\textbf{Motivation.}
As illustrated in Fig.~\ref{fig:mamba_decoder}, given encoded voxel features $\mathbf{F}=\{\mathbf{f}_n\}_{n=1}^{N_v}$ and an instance query set $\mathcal{Q}=\{\mathbf{q}_k\}_{k=1}^{K}$, the decoder iteratively refines queries to predict instance masks.
Unlike standard Transformer decoders that rely on quadratic self-attention among queries and dense cross-attention over all voxel features, we leverage the efficiency of selective state-space models~\cite{gu2024mamba,yao2026lassm} to replace both operations with spatially grounded Mamba, reducing computational complexity while preserving geometric expressiveness.

\vspace{5pt}
\noindent\textbf{Initialization.}
Voxel features and query features are first projected into a shared latent space of dimension $D'$ via lightweight MLP projections $\phi_f$ and $\phi_q$ as:
\begin{equation}
    \mathbf{h}_n = \phi_f(\mathbf{f}_n) \in \mathbb{R}^{D'}, \qquad \mathbf{z}_k^{(0)} = \phi_q(\mathbf{q}_k) \in \mathbb{R}^{D'}.
\end{equation}
Each decoder layer $\ell \in \{0,\dots,L-1\}$ then transforms $\{\mathbf{z}_k^{(\ell)}\}_{k=1}^{K}$ into $\{\mathbf{z}_k^{(\ell+1)}\}_{k=1}^{K}$ through three stages: local feature aggregation, spatial dual-path Mamba, and feed-forward layer.

\vspace{5pt}
\noindent\textbf{Local feature aggregation.}
Each query $k$ is associated with a 3D anchor position $\boldsymbol{\mu}_k \in 
\mathbb{R}^3$, initialized from the corresponding CHM peak or FPS voxel center and updated as the centroid of its predicted mask after each layer.
Rather than attending to all voxels, query $k$ aggregates evidence only from its 
$\kappa$-nearest neighboring voxels in 3D space as:
\begin{equation}
    \mathcal{N}_k = \mathrm{kNN}_\kappa\!\left(\boldsymbol{\mu}_k;\, \{v\,\mathbf{c}_n\}_{n=1}^{N_v}\right).
\end{equation}
Query-dependent attention weights are computed via a scaled dot product as:
\begin{equation}
    \alpha_{kj} = \operatorname{softmax}_{j \in \mathcal{N}_k}\!\left(
    \frac{(\mathbf{W}_k \mathbf{z}_k^{(\ell)})^{\!\top} \cdot\,(\mathbf{W}_a \mathbf{h}_j)}{\sqrt{D'}}\right),
\end{equation}
and local voxel evidence is aggregated via element-wise gating as:
\begin{equation}
    \mathbf{a}_k^{(\ell)} = \sum_{j \in \mathcal{N}_k} \alpha_{kj}
    \left(\mathbf{W}_q \mathbf{z}_k^{(\ell)} \odot \mathbf{W}_v \mathbf{h}_j\right),
\end{equation}
where $\mathbf{W}_k,\mathbf{W}_a,\mathbf{W}_q,\mathbf{W}_v \in \mathbb{R}^{D'\times D'}$ are learnable projections and $\odot$ denotes element-wise multiplication.
The query is updated with a residual connection as:
\begin{equation}
    \tilde{\mathbf{z}}_k^{(\ell)} = \operatorname{LN}\!\left(\mathbf{z}_k^{(\ell)} + \mathbf{W}_o \mathbf{a}_k^{(\ell)}\right),
\end{equation}
with $\mathbf{W}_o \in \mathbb{R}^{D'\times D'}$.
This local aggregation focuses each query on spatially proximate tree evidence, avoiding the noise introduced by attending to irrelevant background voxels.

\vspace{5pt}
\noindent\textbf{Spatial dual-path Mamba.}
To capture inter-instance dependencies, we serialize the updated queries $\tilde{\mathbf{Z}}^{(\ell)}=\big[\tilde{\mathbf{z}}_1^{(\ell)},\dots,\tilde{\mathbf{z}}_K^{(\ell)}\big]$ according to their 3D anchor positions $\{\boldsymbol{\mu}_k\}_{k=1}^{K}$ and process them with a Mamba layer~\cite{gu2024mamba}.
To mitigate the ordering bias inherent in causal sequential scanning, we employ two complementary slab-wise traversals, $\pi_1$ and $\pi_2$, in opposite directions (e.g., bottom-up and top-down).
For each ordering, the query sequence is processed independently by a shared Mamba block as:
\begin{equation}
    \mathbf{Y}^{(\ell,r)} = \pi_r^{-1}\!\left(\operatorname{Mamba}\!\left(\operatorname{LN}\!
    \big(\tilde{\mathbf{Z}}^{(\ell)}\big)_{\pi_r}\right)\right), \quad r \in \{1,2\}.
\end{equation}
The two outputs are then fused via residual averaging as:
\begin{equation}
    \hat{\mathbf{Z}}^{(\ell)} = \operatorname{LN}\!\left(\tilde{\mathbf{Z}}^{(\ell)} +
    \tfrac{1}{2}\!\left(\mathbf{Y}^{(\ell,1)} + \mathbf{Y}^{(\ell,2)}\right)\right).
\end{equation}
A feed-forward network with residual connection then produces the layer output as:
\begin{equation}
    \mathbf{Z}^{(\ell+1)} = \operatorname{LN}\!\left(\hat{\mathbf{Z}}^{(\ell)} + \operatorname{FFN}(\hat{\mathbf{Z}}^{(\ell)})\right).
\end{equation}

\vspace{5pt}
\noindent\textbf{Mask prediction.}
After each decoder layer $\ell$, an objectness score $s$ and an instance mask $\mathbf{M}$ are predicted for every query.
Mask logits are computed as a dot product between the normalized query embedding and projected voxel features as:
\begin{equation}
    \mathbf{M}_{k,n}^{(\ell)} = \operatorname{LN}\!\big(\mathbf{z}_k^{(\ell)}\big)^{\!\top} \cdot \psi(\mathbf{f}_n), \quad
    s_k^{(\ell)} = \sigma\!\left(\mathbf{w}_s^{\!\top} \cdot \mathbf{z}_k^{(\ell)}\right),
\end{equation}
where $\psi:\mathbb{R}^{D}\!\to\!\mathbb{R}^{D'}$ is a learned mask-feature projection, $\mathbf{w}_s\in\mathbb{R}^{D'}$ is the objectness head, and $\sigma$ is the sigmoid function.
Iterative predictions across all $L$ decoder layers provide auxiliary supervision and stabilize training.

\subsection{Training objective}
We train the model end-to-end with a multi-task loss as:
\begin{equation}
\mathcal{L} =
\lambda_{\text{ins}}\,\mathcal{L}_{\text{ins}} \;+\; \lambda_{\text{sem}}\,\mathcal{L}_{\text{sem}} \;+\;
\lambda_{\text{bin}}\,\mathcal{L}_{\text{bin}} \;+\;
\lambda_{\text{dis}}\,\mathcal{L}_{\text{dis}},
\end{equation}
where $\lambda_*$ are scalar weights balancing the loss terms.
Here, $\mathcal{L}_{\text{sem}}$ is the cross-entropy semantic segmentation loss,
$\mathcal{L}_{\text{bin}}$ supervises tree/non-tree classification in the auxiliary branch, and $\mathcal{L}_{\text{dis}}$ is a discriminative embedding loss~\cite{de2017semantic} that enforces intra-instance compactness and inter-instance separation among tree-voxel embeddings $\{\mathbf{e}_n\}$.
The instance loss $\mathcal{L}_{\text{ins}}$ supervises the query decoder via one-to-many matching between predicted and ground-truth masks, and is averaged over all $L$ decoder layers as:
\begin{equation}
\mathcal{L}_{\text{ins}} = \frac{1}{L}\sum_{\ell=1}^{L}\!\left(
\lambda_{\text{cls}}\,\mathcal{L}_{\text{cls}}^{(\ell)} + \lambda_{\text{BCE}}\,\mathcal{L}_{\text{BCE}}^{(\ell)} + \lambda_{\text{Dice}}\,\mathcal{L}_{\text{Dice}}^{(\ell)}
\right),
\end{equation}
where $\mathcal{L}_{\text{cls}}^{(\ell)}$ is the objectness loss on the predicted scores $\{s_k^{(\ell)}\}$,
$\mathcal{L}_{\text{BCE}}^{(\ell)}$ is the Binary Cross-Entropy (BCE) on mask logits $\{\mathbf{M}_k^{(\ell)}\}$,
$\mathcal{L}_{\text{Dice}}^{(\ell)}$ is a Dice loss enforcing spatial overlap with the matched ground-truth masks, and $\lambda_*$ are scalar weights.
This iterative supervision provides auxiliary signals at every decoder layer.

\section{Experiments}
\label{sec:experimental_results}
\subsection{Experimental conditions}
\noindent \textbf{Dataset.}
We evaluate the proposed \textbf{ForestMamba} on the FOR-instanceV2 dataset~\cite{xiang2025forestformer3d}, which comprises seven geographically diverse forest regions across four continents, covering boreal, temperate, and tropical forest types: CULS (Czech Republic), NIBIO (Norway), RMIT (Australia), SCION (New Zealand), TUWIEN (Austria), BlueCat (Czech Republic), and Yuchen (French Guiana). 
The dataset contains 8{,}016 trees in the training set, 1{,}374 in the validation set, and 1{,}744 in the test set, totaling 11{,}134 annotated tree instances. 
Only raw $(x,y,z)$ coordinates are used as input, without color or intensity information.

During training, each plot is cropped into cylindrical patches of radius 16\,m centered at randomly sampled locations. 
Points are voxelized at a resolution of 0.2\,m and subsampled to at most 640\,k points per patch. 
Data augmentation includes random horizontal flipping, global rotation around the vertical axis, and uniform scaling in the range $[0.8, 1.2]$.
During validation, each plot is processed in a single forward-pass on the full point cloud, without cropping.
A sliding cylindrical window of radius 16\,m is applied across the full plot with a stride of 4\,m, resulting in $75\%$ overlap between adjacent windows. 
Predictions from all windows are merged into a global instance map using a best-score-wins strategy, followed by score-based instance merging to resolve remaining overlaps.

\vspace{5pt}
\noindent\textbf{Evaluation metrics.}
We evaluate both instance and semantic segmentation.
\begin{itemize}
    \item \textbf{Instance segmentation.} A predicted instance is considered a True Positive (TP) if its Intersection over Union (IoU) with a ground-truth instance exceeds 0.5. 
    Otherwise, it is a False Positive (FP), and unmatched ground-truth instances are False Negatives (FN). 
    We report Precision (P), Recall (R), and F$_1$ score. 
    We also report Coverage (Cov), defined as the average maximum IoU between each ground-truth instance $\mathcal{M}_g$ and all predicted masks $\hat{\mathcal{M}}_k$, i.e., $\mathrm{Cov} = \frac{1}{|\mathcal{G}|}\sum_{g \in \mathcal{G}} \max_{k}\, \mathrm{IoU}(\hat{\mathcal{M}}_k, \mathcal{M}_g)$.
    \item \textbf{Semantic segmentation.} We report per-class IoU for Ground ($\mathrm{IoU_G}$), Wood ($\mathrm{IoU_W}$), and Leaf ($\mathrm{IoU_L}$), along with mean IoU (mIoU).
\end{itemize}

\begin{table*}[t]
\centering
\caption{
Comparison with baseline methods on instance and semantic segmentation across different forest regions. The best and second-best results are highlighted in \textbf{bold} and \underline{underline}, respectively. 
* indicates results reported in the original paper.
}
\label{tab:region_comparison}
\resizebox{\textwidth}{!}
{%
\begin{tabular}{clcccccccc}
\toprule
\multirowcell{2}[-2pt][c]{{Region}} & \multirowcell{2}[-2pt][l]{{Method}} & \multicolumn{4}{c}{Instance Segmentation [\%]} & \multicolumn{4}{c}{Semantic Segmentation [\%]} \\
\cmidrule(lr){3-6} \cmidrule(lr){7-10}
&  & P & R & F$_1$ & Cov & IoU$_\text{G}$ & IoU$_\text{W}$ & IoU$_\text{L}$ & mIoU \\
\midrule
\multirow{3}{*}{\shortstack{CULS \\ \small{Czech Republic} \\ \small{(20 trees)}}}
 & OneFormer3D~\cite{kolodiazhnyi2024oneformer3d} & 94.7 & 90.0 & 92.3 & 83.3 & 99.8 & 60.8 & 95.7 & 85.4 \\
 & ForestFormer3D~\cite{xiang2025forestformer3d} & \hspace{-5pt}100.0 & 95.0 & 97.4 & 94.4 & 99.8 & 61.2 & 95.7 & 85.6 \\
 & ForestMamba (Ours) & \hspace{-5pt}100.0 & \hspace{-5pt}100.0 & \hspace{-5pt}100.0 & 99.5 & 99.8 & 62.0 & 95.8 & 85.9 \\
\midrule
\multirow{3}{*}{\shortstack{YuChen \\ \small{French Guiana} \\ \small{(24 trees)}}}
 & OneFormer3D~\cite{kolodiazhnyi2024oneformer3d} & 30.8 & 16.7 & 21.6 & 24.3 & 97.1 & 36.5 & 97.4 & 77.0 \\
 & ForestFormer3D~\cite{xiang2025forestformer3d} & 88.9 & 66.7 & 76.2 & 66.6 & 99.7 & 44.8 & 97.7 & 80.7 \\
 & ForestMamba (Ours) & 88.9 & 66.7 & 76.2 & 68.1 & 99.8 & 38.7 & 97.7 & 78.8 \\
\midrule
\multirow{3}{*}{\shortstack{TUWIEN \\ \small{Austria} \\ \small{(35 trees)}}}
 & OneFormer3D~\cite{kolodiazhnyi2024oneformer3d} & 42.0 & 37.1 & 39.4 & 36.4 & 92.9 & 49.5 & 92.0 & 78.1 \\
 & ForestFormer3D~\cite{xiang2025forestformer3d} & 83.9 & 74.3 & 78.8 & 65.9 & 98.5 & 48.8 & 94.4 & 80.6 \\
 & {ForestMamba (Ours)} & 91.7 & 62.9 & 74.6 & 54.8 & 98.8 & 49.1 & 94.4 & 80.7 \\
\midrule
\multirow{3}{*}{\shortstack{SCION \\ \small{New Zealand} \\ \small{(43 trees)}}}
 & OneFormer3D~\cite{kolodiazhnyi2024oneformer3d} & 89.8 & 81.4 & 85.2 & 74.6 & 99.4 & 43.9 & 93.9 & 79.1 \\
 & ForestFormer3D~\cite{xiang2025forestformer3d} & 97.4 & 86.1 & 91.4 & 80.0 & 99.6 & 36.2 & 93.2 & 76.3 \\
 & ForestMamba (Ours) & 96.9 & 79.1 & 87.0 & 77.2 & 99.1 & 48.4 & 94.4 & 80.6 \\
\midrule
\multirow{3}{*}{\shortstack{RMIT \\ \small{Australia} \\ \small{(64 trees)}}}
 & OneFormer3D~\cite{kolodiazhnyi2024oneformer3d} & 83.3 & 62.5 & 71.4 & 58.8 & 93.2 & 47.4 & 86.0 & 75.5 \\
 & ForestFormer3D~\cite{xiang2025forestformer3d} & 80.7 & 78.1 & 79.4 & 70.3 & 95.6 & 46.4 & 90.0 & 77.3 \\
 & ForestMamba (Ours) & 89.3 & 78.1 & 83.3 & 71.4 & 97.7 & 45.3 & 91.5 & 78.2 \\
\midrule
\multirow{3}{*}{\shortstack{BlueCat \\ \small{Czech Republic} \\ \small{(537 trees)}}}
 & OneFormer3D~\cite{kolodiazhnyi2024oneformer3d} & 75.8 & 59.4 & 66.5 & 57.2 & -- & 55.5 & 91.3 & 73.4 \\
 & ForestFormer3D~\cite{xiang2025forestformer3d} & 87.6 & 63.3 & 73.4 & 60.8 & -- & 65.8 & 93.1 & 79.5 \\
 & ForestMamba (Ours) & 85.4 & 67.2 & 75.1 & 63.9 & -- & 64.4 & 92.9 & 78.6 \\
\midrule
\multirow{3}{*}{\shortstack{NIBIO \\ \small{Norway} \\ \small{(1,021 trees)}}}
 & OneFormer3D~\cite{kolodiazhnyi2024oneformer3d} & 87.4 & 73.0 & 79.3 & 68.5 & 96.2 & 52.7 & 94.7 & 81.2 \\
 & ForestFormer3D~\cite{xiang2025forestformer3d} & 93.8 & 80.1 & 86.2 & 74.9 & 95.7 & 52.5 & 95.0 & 81.1 \\
 & ForestMamba (Ours) & 94.3 & 81.5 & 87.2 & 76.0 & 96.1 & 53.4 & 95.1 & 81.6 \\
\midrule
\midrule
\multirow{5}{*}{\shortstack{Weighted \\ Mean}}
 & TreeLearn$^{*}$~\cite{henrich2024treelearn} & 82.0 & 36.6 & 50.6 & 52.2 & -- & -- & -- & -- \\
 & ForAINet$^{*}$~\cite{xiang2024automated} & 86.7 & 65.8 & 72.8 & 63.9 & \underline{96.4} & 55.5 & 94.0 & {82.0} \\
 & OneFormer3D~\cite{kolodiazhnyi2024oneformer3d} & 82.2 & 67.3 & 74.0 & 63.8 & 96.1 & 53.0 & 93.3 & 80.8 \\
 & ForestFormer3D~\cite{xiang2025forestformer3d} & \underline{91.3} & \underline{74.9} & \underline{82.3} & \underline{70.5} & 96.1 & \underline{55.9} & \underline{94.2} & \underline{82.1} \\
& \textbf{ForestMamba (Ours)} & \textbf{91.4} & \textbf{76.6} & \textbf{83.4} & \textbf{71.9} & \textbf{96.5} & \textbf{56.2} & \textbf{94.3} & \textbf{82.3} \\
\bottomrule
\end{tabular}%
}
\vspace{-5pt}
\end{table*}

\vspace{5pt}
\noindent\textbf{Comparison methods.}
We compare ForestMamba against methods from different segmentation paradigms for point clouds in forest environments.
{TreeLearn}~\cite{henrich2024treelearn} takes a bottom-up approach that groups points via flow-based clustering but struggles in dense canopies due to limited global context. 
{ForAINet}~\cite{xiang2024automated} follows a proposal-based pipeline without an end-to-end query formulation. 
{OneFormer3D}~\cite{kolodiazhnyi2024oneformer3d} is a general query-based Transformer method that unifies semantic and instance segmentation through a mask decoder with Hungarian matching~\cite{kuhn1955hungarian}.
{ForestFormer3D}~\cite{xiang2025forestformer3d} is the State-Of-The-Art (SOTA), extending OneFormer3D~\cite{kolodiazhnyi2024oneformer3d} with one-to-many matching and FPS-based query initialization.

\vspace{5pt}
\noindent \textbf{Models \& hyperparameters.}
We use a 5-level sparse encoder as the backbone, Mamba state dimension $d_\text{state}=16$, depthwise convolution width $d_\text{conv}=4$, expansion factor $\text{expand}=1$, and slab thickness $\tau=5$ voxels. 
We adopt a Mamba-based query decoder with $L=6$ layers, model dimension $d_\text{model}=256$, state dimension $d_\text{state}=64$, and feedforward neural network hidden dimension $1,024$ with Gaussian Error Linear Unit (GELU) activation. 
Iterative prediction and mask-guided attention are enabled. 
We set the query budget to $K=300$, with Canopy Height Model (CHM) grid resolutions $r_f=0.3$\,m and $r_c=0.7$\,m, minimum peak separation $d_\text{min}=1.0$\,m, allometric parameters $(\alpha, \beta)=(0.25, 0.5)$, minimum tree height $1.5$\,m, and cylinder pooling radius $r_p=1.5$\,m. 
We train the model for 3{,}000 epochs using AdamW~\cite{loshchilov2019decoupled} ($\text{lr}=10^{-4}$, weight decay $0.05$), with polynomial learning rate decay (power $0.9$, $4.5{\times}10^5$ iterations) and gradient clipping at norm $10$. 
The loss weights are set to $\lambda_{\text{ins}}=1.0$, $\lambda_{\text{sem}}=0.2$, $\lambda_{\text{bin}}=1.0$, and $\lambda_{\text{dis}}=1.0$. 
For the instance loss, we use $(w_{\text{cls}}, w_{\text{BCE}}, w_{\text{Dice}}) = (1.0, 1.0, 0.5)$. 

\subsection{Quantitative results}
\noindent \textbf{Comparison with baselines.}
Table~\ref{tab:region_comparison} reports the performance of the proposed ForestMamba and baseline methods across seven forest regions. 
Overall, ForestMamba achieved the highest weighted mean F$_1$ of 83.4\% and Recall of 76.6\%, surpassing the SOTA method ForestFormer3D~\cite{xiang2025forestformer3d} by 1.1\% and 1.7\%, respectively, while improving Coverage to 71.9\%.
ForestMamba improved Recall while maintaining competitive Precision, indicating more complete detection of tree instances without introducing excessive FPs. 
Across individual regions, ForestMamba showed clear advantages in large-scale, structurally complex forests such as NIBIO and BlueCat, achieving higher F$_1$ and Coverage than prior methods. 
These improvements highlight the effectiveness of structure-aware modeling in handling dense and overlapping canopies. 
In smaller or less complex regions (e.g., CULS), performance was already near saturation across all methods, and ForestMamba remained competitive. 
For more challenging tropical or mixed forests (e.g., YuChen and SCION), the gains were more moderate, reflecting the greater difficulty posed by highly occluded and heterogeneous structures.
For semantic segmentation, ForestMamba achieved competitive or superior mIoU across most regions, indicating better representation of structurally critical components.
These results demonstrate that ForestMamba achieves generalizable performance across diverse forest environments.

\begin{table}[t]
\centering
\caption{Comparison of model complexity. Inference is measured on an NVIDIA RTX A6000 GPU with $\approx$1\,M voxels, averaged over 20 runs.
``Full Net'' denotes the complete end-to-end inference latency of the model.
}
\label{tab:complexity}
\setlength{\tabcolsep}{3pt}
\resizebox{\linewidth}{!}
{%
\begin{tabular}{lllcccccccc}
\toprule
\multirowcell{2}[-2pt][l]{Method} &
\multirowcell{2}[-2pt][l]{Encoder\\Backbone} &
\multirowcell{2}[-2pt][l]{Decoder\\Type}  &
\multicolumn{3}{c}{\# Params \text{[M]}} &
\multicolumn{3}{c}{Inference \text{[ms]}} &
\multirowcell{2}[-2pt][c]{Peak GPU \\ \text{[MB]}} \\
\cmidrule(lr){4-6} \cmidrule(lr){7-9}
& & & Enc. & Dec. & Total & Enc. & Dec. & Full Net & \\
\midrule

OneFormer3D~\cite{kolodiazhnyi2024oneformer3d}    
& SpConv & Transf. 
& \textbf{10.96} & 6.61 & 17.57 
& \textbf{193.7} & 1,841.0 & 2,069.6  
& 28,420  \\

ForestFormer3D~\cite{xiang2025forestformer3d} 
& SpConv & Transf. 
& \textbf{10.96} & 6.54 & 17.50 
& 196.7 & 1,835.9 & 2,047.6
& 28,413  \\

\textbf{ForestMamba (Ours)} 
& SpSSM & SSM 
& 11.16 & \textbf{6.12} & \textbf{17.28} 
& 246.5 & \hspace{5pt} \textbf{435.7} & \hspace{5pt} \textbf{684.2} 
& \textbf{12,130}  \\

\midrule
Change vs.\ SOTA
&  & &
{$\uparrow$1.02$\times$} & \textbf{$\downarrow$1.07$\times$} & \textbf{$\downarrow$1.01$\times$}
& {$\uparrow$1.27$\times$} & \textbf{$\downarrow$4.21$\times$} & \textbf{$\downarrow$2.99$\times$}
& \textbf{$\downarrow$2.34$\times$} \\

\bottomrule
\end{tabular}%
}
\end{table}

\vspace{5pt}
\noindent \textbf{Model complexity analysis.}
Table~\ref{tab:complexity} compares the computational complexity of ForestMamba with representative Transformer-based methods, including OneFormer3D~\cite{kolodiazhnyi2024oneformer3d} and ForestFormer3D~\cite{xiang2025forestformer3d}. 
All models are evaluated under the same setting for a fair comparison.
In terms of model size, ForestMamba achieved a slightly lower total parameter count compared to the baselines, while demonstrating substantial improvements in computational efficiency.
The decoder inference time was reduced by 4.21$\times$, resulting in an overall inference speedup of 2.99$\times$. 
This gain is primarily due to the linear-time complexity of state-space models, in contrast to the quadratic complexity of Transformer attention.
Although the encoder runtime is higher, the overall inference remains significantly faster, driven by the highly efficient decoder.
In addition, ForestMamba substantially reduced memory consumption, requiring 2.34$\times$ less GPU memory.

\vspace{5pt}
\noindent \textbf{Ablation studies.}
Table~\ref{tab:ablation} presents component-wise ablation studies validating each design decision in ForestMamba.

\vspace{2pt}
\noindent\textit{(A) Encoder.}
Reducing the use of Mamba blocks in the encoder decreased Recall by $7.1\%$ and F$_1$ by $4.8\%$, confirming that Mamba effectively captures long-range vertical canopy structure that standard sparse U-Net architectures~\cite{ronneberger2015u} cannot model.

\vspace{2pt}
\noindent\textit{(B) Query initialization.}
We observed that FPS-only queries achieved a comparable F$_1$ score as ablation (A), indicating that learning-driven seeds fail in densely overlapping canopies. 
Single-scale CHM yielded the highest Precision at the expense of Recall, as a fixed allometric window tends to miss suppressed trees. 
The introduction of multi-scale CHM improved Recall by $3.0\%$ and Coverage by $2.0\%$, demonstrating the benefit of adaptive radius scaling across different tree heights.

\begin{table*}[t]
\centering
\caption{
Ablation study of the proposed ForestMamba.
Each component is evaluated by removing or modifying one design at a time.
The best and second-best results are highlighted in \textbf{bold} and \underline{underline}, respectively.
}
\label{tab:ablation}
\resizebox{0.92\linewidth}{!}{
\begin{tabular}{lcccccccc}
\toprule
\multirowcell{2}[-2pt][l]{Setting}
  & \multicolumn{4}{c}{Instance Segmentation [\%]}
  & \multicolumn{4}{c}{Semantic Segmentation [\%]} \\
\cmidrule(lr){2-5} \cmidrule(lr){6-9}
& P & R & F$_1$ & Cov
& IoU$_\text{G}$ & IoU$_\text{W}$ & IoU$_\text{L}$ & mIoU \\
\midrule
\textbf{ForestMamba} & \underline{91.4} & \textbf{76.6} & \textbf{83.4} & \textbf{71.9} & \textbf{96.5} & {56.2} & {94.3} & 82.3  \\
\addlinespace[3pt]
\multicolumn{9}{l}{\textit{(A) Encoder}} \\
\quad w/o Mamba block
  & 90.3 & 69.5 & 78.6 & 67.1 & 95.9 & 57.6 & \textbf{94.5 }& 82.7 \\

\addlinespace[3pt]
\multicolumn{9}{l}{\textit{(B) Query initialization}} \\
\quad FPS query (w/o CHM)
  & 90.3 & 69.4 & 78.5 & 66.8 & 95.9 & 57.8 & \underline{94.4} & 82.7 \\
\quad Single-scale CHM 
  &  \textbf{92.1} & \underline{73.6} & \underline{81.8} & \underline{69.9} & 96.0 & 56.6 & 94.2 & 82.3 \\
\addlinespace[3pt]
\multicolumn{9}{l}{\textit{(C) Decoder}} \\
\quad w/o $k$NN aggregation
  & 90.2 & 67.1 & 78.0 & 65.0 & \underline{96.4} & \underline{59.2} & \underline{94.4} & \textbf{83.3} \\
\quad Single-path Mamba 
  & 91.3 & 66.1 & 76.7 & 64.4 & 95.7 & \textbf{59.5} & \underline{94.4} & \underline{83.2} \\
\addlinespace[3pt]
\multicolumn{9}{l}{\textit{(D) Training strategy}} \\
\quad One-to-Many $\rightarrow$ One-to-One
  & 85.8 & 62.4 & 72.3 & 61.6 & 87.3 & 56.9 & 92.5 & 78.9 \\
\bottomrule
\end{tabular}
}
\vspace{-5pt}
\end{table*}

\vspace{2pt}
\noindent\textit{(C) Decoder.}
Removing $k$NN aggregation dropped Recall by $9.5\%$ and F$_1$ by $5.4\%$, confirming its role in resolving ambiguous crown boundaries.
We found that single-path Mamba caused the steepest Recall drop $10.5\%$ due to the causal directional bias of State Space Model (SSM), where queries early in the scan receive no context from later spatial positions.
Dual-path scanning resolves this by averaging complementary ascending and descending orderings through a shared Mamba block, providing every query with bidirectional spatial context at no additional parameter cost.

\vspace{2pt}
\noindent\textit{(D) Training strategy.}
Replacing one-to-many with one-to-one Hungarian matching~\cite{kuhn1955hungarian} caused the largest degradation, with F$_1$ dropping $11.1\%$ and mIoU dropping $3.4\%$.
One-to-many matching assigns positive supervision to all queries overlapping a ground-truth crown, whereas the Hungarian constraint permits only one match per instance, leaving well-positioned queries unsupervised in crowded canopy regions.

\subsection{Qualitative results}

\begin{figure}[t]
  \centering
  \includegraphics[width=\linewidth]{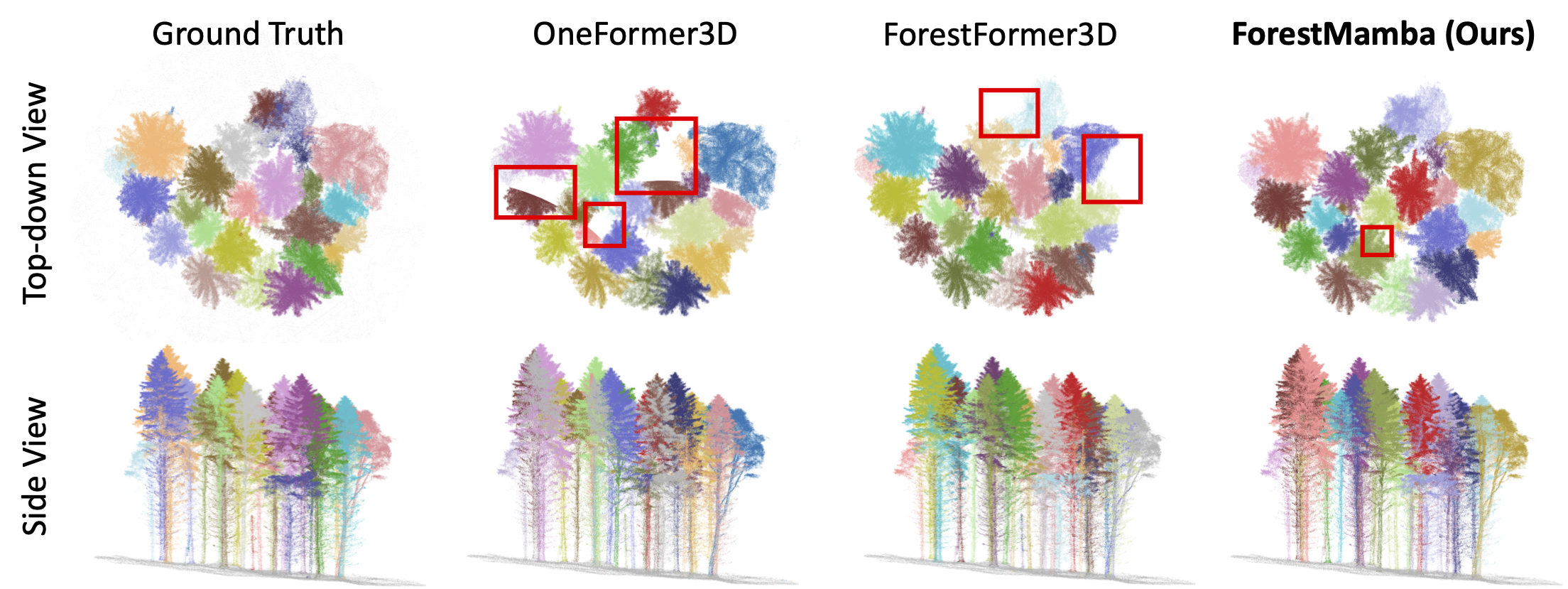}\\[2pt]
  {\small (a) NIBIO (Norway).}\\[6pt]
  \includegraphics[width=\linewidth]{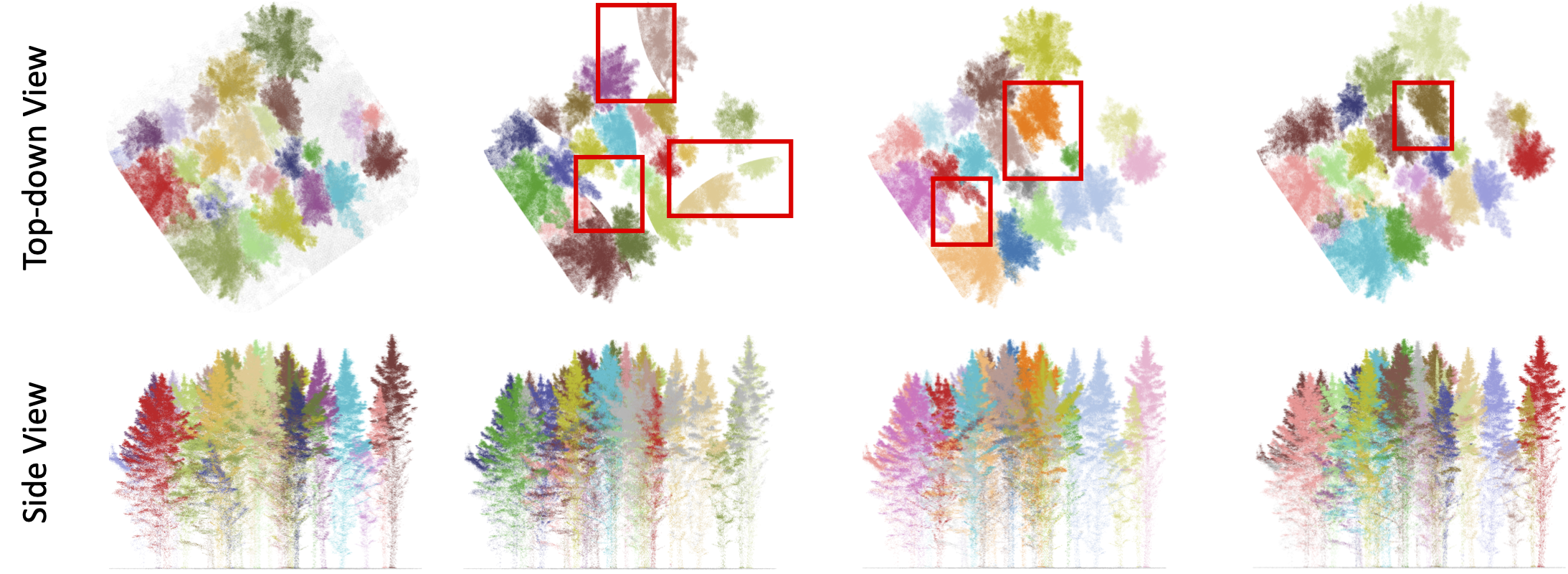}\\[2pt]
  {\small (b) SCION (New Zealand).}
  \vspace{2pt}
  \caption{\textbf{Qualitative comparison of instance segmentation}.
  Top-down and side views are shown for each scene, where ground and unlabeled points are omitted in the top-down view.
  Red boxes mark missing or fragmented predictions. 
  Coloring is random per instance.}
  \label{fig:qualitative}
  \vspace{-5pt}
\end{figure}

As shown in Fig.~\ref{fig:qualitative}, the proposed ForestMamba produced more complete and coherent tree instances compared to baseline methods. 
In both NIBIO and SCION scenes, OneFormer3D~\cite{kolodiazhnyi2024oneformer3d} and ForestFormer3D~\cite{xiang2025forestformer3d} frequently exhibited missing or fragmented predictions, particularly in dense canopy regions and for partially occluded trees (highlighted by red boxes). 
These errors often correspond to weak or ambiguous features, where generic query initialization and attention-based context modeling struggle to separate nearby instances. 
In contrast, ForestMamba better preserved instance continuity and accurately delineates individual trees. 
This improvement was especially evident in top-down views, where ForestMamba recovered missing trees and reduces fragmentation, and in side views, where vertical structures such as stems and crowns were more consistently segmented.

\section{Conclusion}
\label{sec:conclusion}

In this paper, we proposed \textbf{ForestMamba}, a structure-aware framework for joint semantic and instance segmentation of 3D forest point clouds aiming to advance efficiency and accuracy of forest segmentation in close-range LiDAR data.
For this, we introduced three key components: (1) Sparse encoder with vertical-priority slab serialization for efficient long-range context modeling, (2) Geometry-guided query initialization strategy based on multi-scale Canopy Height Model (CHM) peak detection with Farthest Point Sampling (FPS) supplementation, and (3) Mamba-based query decoder with local $k$NN aggregation and spatial dual-path scanning. 
Experiments across seven diverse forest datasets demonstrated that ForestMamba outperformed existing baselines overall in both instance and semantic segmentation, while achieving approximately 3$\times$ faster inference and 2.3$\times$ lower GPU memory than Transformer-based methods.
The source code is publicly available at \url{https://github.com/thanhhff/ForestMamba}.

\vspace{3pt}
\noindent\textbf{Limitations.}
ForestMamba relies on the auxiliary branch for reliable tree-voxel prediction to construct CHM, which may degrade in dense, multi-layered forests.
Additionally, the fixed allometric parameters may not generalize optimally across all forest types.

\vspace{3pt}
\noindent\textbf{Future work.}
We plan to explore adaptive priors beyond CHMs that facilitate the segmentation of understory trees and extend ForestMamba to downstream ecological tasks such as biomass estimation and species classification.

\vspace{5pt}
\noindent \textbf{Acknowledgement.}
This study was supported by the Deutsche Forschungsgemeinschaft (DFG) through Germany's Excellence Strategy (Future Forests–EXC-3127-533786343) and by the Eva Mayr-Stihl Foundation.

\bibliography{egbib}

\clearpage
\appendix

\section{Full Loss Function Specification}
\label{sec:loss_detail}

We train the model end-to-end with a multi-task loss:
\begin{equation}
\mathcal{L} =
\lambda_{\text{ins}}\,\mathcal{L}_{\text{ins}} \;+\; \lambda_{\text{sem}}\,\mathcal{L}_{\text{sem}} \;+\;
\lambda_{\text{bin}}\,\mathcal{L}_{\text{bin}} \;+\;
\lambda_{\text{dis}}\,\mathcal{L}_{\text{dis}},
\end{equation}
where $\lambda_*$ are scalar weights balancing the loss terms.
Here, $\mathcal{L}_{\text{sem}}$ is the cross-entropy semantic segmentation loss,
$\mathcal{L}_{\text{bin}}$ supervises tree/non-tree classification in the auxiliary branch, and $\mathcal{L}_{\text{dis}}$ is a discriminative embedding loss~\cite{de2017semantic} that enforces intra-instance compactness and inter-instance separation among tree-voxel embeddings $\{\mathbf{e}_n\}$.
The instance loss $\mathcal{L}_{\text{ins}}$ supervises the query decoder via one-to-many matching between predicted and ground-truth masks, and is averaged over all $L$ decoder layers:
\begin{equation}
\mathcal{L}_{\text{ins}} = \frac{1}{L}\sum_{\ell=1}^{L}\!\left(
\lambda_{\text{cls}}\,\mathcal{L}_{\text{cls}}^{(\ell)} + \lambda_{\text{bce}}\,\mathcal{L}_{\text{bce}}^{(\ell)} + \lambda_{\text{dice}}\,\mathcal{L}_{\text{dice}}^{(\ell)}
\right),
\end{equation}
where $\mathcal{L}_{\text{cls}}^{(\ell)}$ is the objectness loss on the predicted scores $\{s_k^{(\ell)}\}$,
$\mathcal{L}_{\text{bce}}^{(\ell)}$ is the binary cross-entropy on mask logits $\{\mathbf{M}_k^{(\ell)}\}$,
$\mathcal{L}_{\text{dice}}^{(\ell)}$ is a Dice loss enforcing spatial overlap with the matched ground-truth masks, and $\lambda_*$ are scalar weights as before.
This iterative supervision provides auxiliary signals at every decoder layer.
\subsection{Semantic Segmentation Loss $\mathcal{L}_{\text{sem}}$}
\label{sec:loss_sem}

Let $C$ denote the number of semantic classes (ground, wood, leaf).
For each voxel $n$, let $\hat{\mathbf{p}}_n^{\text{sem}} \in \mathbb{R}^C$ be the predicted class log-probabilities and $y_n^{\text{sem}} \in \{1,\ldots,C\}$ the ground-truth label.
The semantic loss is the standard cross-entropy:
\begin{equation}
  \mathcal{L}_{\text{sem}}
  = -\frac{1}{N_v}\sum_{n=1}^{N_v}
    \log\!\left(\operatorname{softmax}(\hat{\mathbf{p}}_n^{\text{sem}})_{y_n^{\text{sem}}}\right).
\end{equation}

\subsection{Binary Tree/Background Loss $\mathcal{L}_{\text{bin}}$}
\label{sec:loss_bin}

The auxiliary branch predicts a tree/non-tree binary logit $b_n \in \mathbb{R}$ per voxel.
Let $t_n \in \{0,1\}$ be the binary ground-truth label ($t_n=1$ if voxel $n$ belongs to any tree instance).
The binary classification loss is:
\begin{equation}
  \mathcal{L}_{\text{bin}}
  = -\frac{1}{N_v}\sum_{n=1}^{N_v}
    \Big[t_n \log\sigma(b_n)
    + (1-t_n)\log(1-\sigma(b_n))\Big],
\end{equation}
where $\sigma(\cdot)$ denotes the sigmoid function.

\subsection{Discriminative Embedding Loss $\mathcal{L}_{\text{dis}}$}
\label{sec:loss_dis}

Following~\cite{de2017semantic}, the discriminative loss is applied to tree-voxel embeddings
$\{\mathbf{e}_n \in \mathbb{R}^{D_e}\}_{n \in \mathcal{T}}$ to encourage intra-instance compactness and inter-instance separation.
Let $M$ be the number of tree instances, $C_m \subseteq \mathcal{T}$ the set of voxels belonging to instance $m$, and
\begin{equation}
  \boldsymbol{\mu}_m = \frac{1}{|C_m|}\sum_{n \in C_m} \mathbf{e}_n
\end{equation}
the mean embedding (centroid) of instance $m$.
The loss consists of three terms:
\begin{align}
  \mathcal{L}_{\text{var}}
  &= \frac{1}{M}\sum_{m=1}^{M}
    \frac{1}{|C_m|}\sum_{n \in C_m}
    \Big[\|\boldsymbol{\mu}_m - \mathbf{e}_n\|_2 - \delta_v\Big]_+^2,
  \label{eq:var}\\
  \mathcal{L}_{\text{dist}}
  &= \frac{1}{M(M-1)}
    \sum_{\substack{a,b=1\\a \neq b}}^{M}
    \Big[2\delta_d - \|\boldsymbol{\mu}_a - \boldsymbol{\mu}_b\|_2\Big]_+^2,
  \label{eq:dist}\\
  \mathcal{L}_{\text{reg}}
  &= \frac{1}{M}\sum_{m=1}^{M}\|\boldsymbol{\mu}_m\|_2,
  \label{eq:reg}
\end{align}
where $[\cdot]_+ = \max(0,\cdot)$, $\delta_v$ is the intra-cluster pull margin,
and $\delta_d$ is the inter-cluster push margin.
We set $\delta_v = 0.5$, $\delta_d = 1.5$, and weight the three terms equally:
\begin{equation}
  \mathcal{L}_{\text{dis}}
  = \mathcal{L}_{\text{var}} + \mathcal{L}_{\text{dist}} + \mathcal{L}_{\text{reg}}.
\end{equation}
The variance term (Equation~\ref{eq:var}) pulls same-instance embeddings toward their centroid,
the distance term (Equation~\ref{eq:dist}) pushes different-instance centroids apart,
and the regularization term (Equation~\ref{eq:reg}) prevents embedding collapse.

\subsection{Instance Loss $\mathcal{L}_{\text{ins}}$ and One-to-Many Matching}
\label{sec:loss_ins}

\paragraph{One-to-many matching.}
At each decoder layer $\ell$, we first binarize the predicted mask for query $k$ as:
\begin{equation}
  \hat{\mathcal{M}}_k^{(\ell)}
  = \bigl\{n : \sigma\!\bigl(\mathbf{M}_{k,n}^{(\ell)}\bigr) > 0.5\bigr\}.
\end{equation}
The positive query set $\mathcal{P}^+$ is defined by IoU overlap with any ground-truth tree instance:
\begin{equation}
  \mathcal{P}^+ = \Bigl\{k \;\Big|\; \exists\,m \in \{1,\ldots,M\},\;
  \operatorname{IoU}\!\bigl(\hat{\mathcal{M}}_k^{(\ell)},\,\mathcal{M}_m\bigr) \geq \theta\Bigr\},
  \label{eq:pos_set}
\end{equation}
with $\theta = 0.5$.
Note that multiple queries can be simultaneously matched to the same ground-truth instance,
unlike Hungarian one-to-one matching.
Each positive query $k \in \mathcal{P}^+$ is assigned to its best-matching instance:
\begin{equation}
  m^*(k) = \arg\max_{m}\,\operatorname{IoU}\!\bigl(\hat{\mathcal{M}}_k^{(\ell)},\,\mathcal{M}_m\bigr),
\end{equation}
yielding the per-voxel binary target
$y_{k,n}^* = \mathbf{1}[n \in \mathcal{M}_{m^*(k)}]$.
Queries in $\mathcal{P}^- = \{1,\ldots,K\}\setminus\mathcal{P}^+$ are treated as negative.

\paragraph{Objectness loss $\mathcal{L}_{\text{cls}}^{(\ell)}$.}
Each query receives a binary objectness target $q_k^* = \mathbf{1}[k \in \mathcal{P}^+]$.
The objectness score $s_k^{(\ell)} = \sigma(\mathbf{w}_s^\top \mathbf{z}_k^{(\ell)})$ is supervised by:
\begin{equation}
  \mathcal{L}_{\text{cls}}^{(\ell)}
  = -\frac{1}{K}\sum_{k=1}^{K}
  \Bigl[q_k^*\log s_k^{(\ell)}
  + (1-q_k^*)\log\!\bigl(1-s_k^{(\ell)}\bigr)\Bigr].
\end{equation}

\paragraph{Binary cross-entropy mask loss $\mathcal{L}_{\text{bce}}^{(\ell)}$.}
Applied only to positive queries:
\begin{equation}
  \mathcal{L}_{\text{bce}}^{(\ell)}
  = -\frac{1}{|\mathcal{P}^+|}\sum_{k \in \mathcal{P}^+}
  \frac{1}{N_v}\sum_{n=1}^{N_v}
  \Bigl[y_{k,n}^*\log\hat{p}_{k,n}
  + (1-y_{k,n}^*)\log(1-\hat{p}_{k,n})\Bigr],
\end{equation}
where $\hat{p}_{k,n} = \sigma(\mathbf{M}_{k,n}^{(\ell)})$ is the predicted probability that voxel $n$ belongs to the instance of query $k$.

\paragraph{Dice loss $\mathcal{L}_{\text{dice}}^{(\ell)}$.}
The Dice loss~\cite{milletari2016vnet} measures spatial overlap between the soft mask and the binary target:
\begin{equation}
  \mathcal{L}_{\text{dice}}^{(\ell)}
  = \frac{1}{|\mathcal{P}^+|}\sum_{k \in \mathcal{P}^+}
  \left(1 -
    \frac{2\displaystyle\sum_{n=1}^{N_v} \hat{p}_{k,n}\,y_{k,n}^* + \epsilon}
         {\displaystyle\sum_{n=1}^{N_v} \hat{p}_{k,n}
          + \sum_{n=1}^{N_v} y_{k,n}^* + \epsilon}
  \right),
\end{equation}
with numerical stabilizer $\epsilon = 10^{-6}$.
The Dice loss is scale-invariant with respect to the number of foreground voxels,
compensating for the severe foreground/background imbalance in large forest scenes.

\paragraph{Combined instance loss.}
Summing across decoder layers:
\begin{equation}
  \mathcal{L}_{\text{ins}}
  = \frac{1}{L}\sum_{\ell=1}^{L}
  \Bigl(
    \lambda_{\text{cls}}\,\mathcal{L}_{\text{cls}}^{(\ell)}
    + \lambda_{\text{bce}}\,\mathcal{L}_{\text{bce}}^{(\ell)}
    + \lambda_{\text{dice}}\,\mathcal{L}_{\text{dice}}^{(\ell)}
  \Bigr).
\end{equation}

\section{Dataset}
\label{sec:algo}
\begin{table}[t]
\centering
\caption{Summary of the FOR-instanceV2 dataset used in this work. 
The dataset spans seven geographically diverse forest regions across four continents and includes boreal, temperate, and tropical forest types acquired using different LiDAR platforms. 
\# Train, \# Val, and \# Test denote the number of annotated tree instances in each split for every region.
}
\label{tab:dataset}
\resizebox{\linewidth}{!}
{%
\begin{tabular}{llllrrr}
\toprule
Region & Country & Forest Type & Sensor & \# Train & \# Val & \# Test \\
\midrule
CULS    & Czech Republic & Temperate coniferous & ULS & 6 & 21 & 20 \\
NIBIO   & Norway & Boreal coniferous & ULS / MLS & 2401 & 572 & 1021 \\
RMIT    & Australia & Dry eucalypt forest & ULS & 159 & 0 & 64 \\
SCION   & New Zealand & Temperate coniferous & ULS & 69 & 23 & 43 \\
TUWIEN  & Austria & Temperate deciduous & ULS & 115 & 0 & 35 \\
BlueCat & Czech Republic & Temperate mixed forest & TLS & 5032 & 735 & 537 \\
Yuchen  & French Guiana & Tropical forest & ULS & 234 & 23 & 24 \\
\midrule
Total &  &  &  & 8016 & 1374 & 1744 \\
\bottomrule
\end{tabular}%
}
\end{table}

\noindent We evaluate the proposed ForestMamba on the FOR-instanceV2 dataset~\cite{xiang2025forestformer3d}, which contains close-range forest point clouds collected from seven geographically diverse regions across four continents, as summarized in Table~\ref{tab:dataset}. 
The dataset includes temperate coniferous forests from CULS (Czech Republic) and SCION (New Zealand), boreal coniferous forests from NIBIO (Norway), temperate deciduous forests from TUWIEN (Austria), mixed temperate forests from BlueCat (Czech Republic), dry eucalypt forests from RMIT (Australia), and tropical forests from Yuchen (French Guiana). 
As shown in Table~\ref{tab:dataset}, the point clouds are acquired using Unmanned Laser Scanning (ULS), Terrestrial Laser Scanning (TLS), and Mobile Laser Scanning (MLS) systems, providing substantial variations in forest structure, density, terrain, and acquisition geometry. 
Overall, the dataset contains 8{,}016 annotated tree instances for training, 1{,}374 for validation, and 1{,}744 for testing, totaling 11{,}134 trees. 
Each point is annotated with both semantic labels (ground, wood, and leaf) and individual tree instance IDs. 
Only raw XYZ coordinates are used as input, without color or intensity information.

\section{Algorithm Pseudo-code}
\label{sec:algo}

\subsection{Multi-scale CHM Peak Detection}

Algorithm~\ref{alg:chm} details the complete CHM peak detection and cross-scale NMS procedure
described in Section~\ref{sec:query_init} of the main paper.
The procedure runs in three stages.
\begin{itemize}
    \item First, for each resolution $r \in \mathcal{R}$, a 2D height grid $H^{(r)}$ is built by projecting tree voxels onto the horizontal plane and retaining the maximum height per cell.
    \item Second, local maxima are identified by comparing each cell against its neighborhood of radius $\rho$, where $\rho$ grows allometrically with tree height so that taller trees suppress competition over a larger area.
    This height-adaptive suppression avoids the over-detection of dominant trees that fixed-radius methods suffer from in structurally heterogeneous stands.
    \item Third, candidate peaks from all resolutions are merged into a single set and deduplicated via greedy height-sorted NMS: peaks are accepted in descending order of height, and a new peak is retained only if it lies at least $d_{\min}$ meters from every already-accepted peak.
    This cross-scale deduplication resolves the ambiguity that arises when the same treetop is detected at multiple resolutions, ensuring that the final peak set $\{\mathbf{m}_k\}$ contains at most $K$ spatially distinct candidates ranked by canopy height.
\end{itemize}

\begin{algorithm}[t]
\caption{Multi-scale CHM Peak Detection}
\label{alg:chm}
\begin{algorithmic}[1]
\Input{Tree voxels $\mathcal{T}$; voxel size $v$; scales $\mathcal{R}=\{r_f, r_c\}$;
       allometric params $(\alpha,\beta)$; min.\ height $h_{\min}$;
       min.\ separation $d_{\min}$; query budget $K$.}
\Output{CHM-guided query positions $\{\mathbf{m}_k\}_{k=1}^{K_{\mathrm{CHM}}}$ and heights $\{H_k\}$.}
\State $\mathcal{C}_{\mathrm{all}} \gets \emptyset$
\For{each resolution $r \in \mathcal{R}$}
  \State Build 2D grid $H^{(r)}$ with cell size $r$:
  \Statex \quad $H^{(r)}(a,b) \gets \max\{h_n \mid n \in \mathcal{T},\;\lfloor\mathbf{u}_n/r\rfloor=(a,b)\}$
  \For{each cell $(a,b)$ with $H^{(r)}(a,b) \geq h_{\min}$}
    \State $h \gets H^{(r)}(a,b)$; \quad $\rho \gets \lceil \alpha\,h^{\beta}/r \rceil$
    \If{$H^{(r)}(a,b) \geq H^{(r)}(a',b')$ for all $(a',b') \in \mathcal{N}_\rho(a,b)$}
      \State $\mathbf{m} \gets r \cdot (a,b) + r/2$ \Comment{Cell centre in metric coords}
      \State $\mathcal{C}_{\mathrm{all}} \gets \mathcal{C}_{\mathrm{all}} \cup \{(\mathbf{m},\, h)\}$
    \EndIf
  \EndFor
\EndFor
\State Sort $\mathcal{C}_{\mathrm{all}}$ by height $h$ in \textbf{descending} order
\State $\mathcal{P}_{\mathrm{kept}} \gets \emptyset$
\For{each $(\mathbf{m},h)$ in sorted $\mathcal{C}_{\mathrm{all}}$}
  \If{$\|\mathbf{m} - \mathbf{m}'\|_2 \geq d_{\min}$ for all $(\mathbf{m}',h') \in \mathcal{P}_{\mathrm{kept}}$}
    \State $\mathcal{P}_{\mathrm{kept}} \gets \mathcal{P}_{\mathrm{kept}} \cup \{(\mathbf{m},h)\}$
    \If{$|\mathcal{P}_{\mathrm{kept}}| = K$}
      \State \textbf{break}
    \EndIf
  \EndIf
\EndFor
\State \Return $\{\mathbf{m}_k\}_{k=1}^{K_{\mathrm{CHM}}} \gets \mathcal{P}_{\mathrm{kept}}$
\end{algorithmic}
\end{algorithm}

\subsection{Geometry-guided Query Initialization}

Algorithm~\ref{alg:query_init} describes the full query initialization pipeline,
combining CHM-guided seeds with FPS supplementation.
The algorithm proceeds in two complementary stages.
\begin{itemize}
    \item In the first stage, each retained CHM peak $\mathbf{m}_k$ is converted into an instance query by aggregating encoder features from all tree voxels within a cylinder of radius $r_p$ centered at the peak's horizontal position.
    This cylinder pooling produces a query that summarizes the local crown context around the detected treetop, rather than relying on a single voxel feature that may be noisy or unrepresentative.
    The 3D anchor $\boldsymbol{\mu}_k$ is initialized from the peak's world coordinates and CHM height, providing a spatially grounded starting position for the decoder's kNN aggregation.

    \item In the second stage, if fewer than $K$ CHM peaks are available (i.e.\ $K_{\mathrm{CHM}} < K$), the remaining $K - K_{\mathrm{CHM}}$ queries are filled by FPS over the discriminative embeddings $\{\mathbf{e}_n\}$ of tree voxels.
    FPS is seeded deterministically from the embedding farthest from the mean, then iteratively selects the embedding most distant from all previously selected ones, maximizing coverage of the embedding space.
    These FPS queries are designed to complement the CHM seeds by targeting understory trees and occluded crowns that lack clear canopy peaks and would otherwise be missed.
    The two query sets are concatenated to form the final query pool $\mathcal{Q}$ of size exactly $K$, which is passed to the Mamba-based decoder for iterative refinement.

\end{itemize}

\begin{algorithm}[t]
\caption{Geometry-guided Query Initialization}
\label{alg:query_init}
\begin{algorithmic}[1]
\Input{Encoder features $\mathbf{F}=\{\mathbf{f}_n\}$; discriminative embeddings $\{\mathbf{e}_n\}$;
       auxiliary binary predictions $\{b_n\}$; voxel coordinates $\{v\,\mathbf{c}_n\}$;
       cylinder radius $r_p$; query budget $K$.}
\Output{Query feature set $\mathcal{Q}$; anchor positions $\{\boldsymbol{\mu}_k\}$.}
\State $\mathcal{T} \gets \{n : \sigma(b_n) > 0.5\}$ \Comment{Predicted tree voxels}
\State $\{\mathbf{m}_k, H_k\}_{k=1}^{K_{\mathrm{CHM}}} \gets \textsc{ChmPeakDetect}(\mathcal{T})$
  \Comment{Algorithm~\ref{alg:chm}}
\State \emph{// CHM-guided queries via cylinder pooling}
\For{$k = 1$ to $K_{\mathrm{CHM}}$}
  \State $\Omega_k \gets \{n \in \mathcal{T} : \|v\,\mathbf{u}_n - \mathbf{m}_k\|_2 \leq r_p\}$
  \State $\mathbf{q}_k^{\mathrm{CHM}} \gets \frac{1}{|\Omega_k|}\sum_{n \in \Omega_k}\mathbf{f}_n$
  \State $\boldsymbol{\mu}_k^{\mathrm{CHM}} \gets (\mathbf{m}_k,\; H_k)$
    \Comment{3D anchor: horizontal peak + CHM height}
\EndFor
\State \emph{// FPS supplementation for remaining budget}
\State $K_{\mathrm{FPS}} \gets K - K_{\mathrm{CHM}}$
\If{$K_{\mathrm{FPS}} > 0$}
  \State $\mathcal{E}_{\mathrm{tree}} \gets \{\mathbf{e}_n \mid n \in \mathcal{T}\}$
  \State $\bar{\mathbf{e}} \gets \frac{1}{|\mathcal{T}|}\sum_{n \in \mathcal{T}}\mathbf{e}_n$
    \Comment{Centroid for deterministic seeding}
  \State $\mathcal{S} \gets \emptyset$;
    $i_1 \gets \arg\max_{j \in \mathcal{T}}\|\mathbf{e}_j - \bar{\mathbf{e}}\|_2$;
    $\mathcal{S} \gets \{i_1\}$
  \For{$t = 2$ to $K_{\mathrm{FPS}}$}
    \State $i_t \gets \arg\max_{j \in \mathcal{T} \setminus \mathcal{S}}
      \min_{k \in \mathcal{S}}\|\mathbf{e}_j - \mathbf{e}_k\|_2$
    \State $\mathcal{S} \gets \mathcal{S} \cup \{i_t\}$
  \EndFor
  \State $\{\mathbf{q}_t^{\mathrm{FPS}}\}_{t=1}^{K_{\mathrm{FPS}}}
    \gets \{\mathbf{f}_{i_t}\}_{t=1}^{K_{\mathrm{FPS}}}$
  \State $\{\boldsymbol{\mu}_t^{\mathrm{FPS}}\}_{t=1}^{K_{\mathrm{FPS}}}
    \gets \{v\,\mathbf{c}_{i_t}\}_{t=1}^{K_{\mathrm{FPS}}}$
\EndIf
\State $\mathcal{Q} \gets
  \{\mathbf{q}_k^{\mathrm{CHM}}\}_{k=1}^{K_{\mathrm{CHM}}} \cup
  \{\mathbf{q}_t^{\mathrm{FPS}}\}_{t=1}^{K_{\mathrm{FPS}}}$
\State $\{\boldsymbol{\mu}_k\} \gets
  \{\boldsymbol{\mu}_k^{\mathrm{CHM}}\} \cup \{\boldsymbol{\mu}_t^{\mathrm{FPS}}\}$
\State \Return $\mathcal{Q},\;\{\boldsymbol{\mu}_k\}$
\end{algorithmic}
\end{algorithm}

\section{Encoder Architecture Details}
\label{sec:enc_detail}

Table~\ref{tab:enc_arch} summarizes the stage-by-stage configuration of the Sparse Mamba encoder.
The encoder follows a 5-level UNet structure.
At each stage, two sparse convolutional residual blocks are applied first, followed by one Sparse Mamba block.
Downsampling uses a strided sparse convolution (stride 2 in $x,y$; stride 1 in $z$) to preserve vertical resolution.
The UNet decoder applies symmetric transposed sparse convolutions with skip connections.

\begin{table}[th]
\centering
\caption{Sparse Mamba encoder stage configuration.}
\label{tab:enc_arch}
\setlength{\tabcolsep}{6pt}
\begin{tabular}{ccccc}
\toprule
Stage & Channels & SpConv Blocks & Mamba Block & Downsample \\
\midrule
1 & 32  & 2 & $\checkmark$ & $\checkmark$ \\
2 & 64  & 2 & $\checkmark$ & $\checkmark$ \\
3 & 128 & 2 & $\checkmark$ & $\checkmark$ \\
4 & 256 & 2 & $\checkmark$ & $\checkmark$ \\
5 & 256 & 2 & $\checkmark$ & $\times$     \\
\bottomrule
\end{tabular}
\end{table}

\vspace{5pt}
\noindent\textbf{Slab serialization.}
For each occupied voxel $n$ at encoder stage $\ell$, the slab index is
$s_n = \lfloor c_n^z / \tau \rfloor$ with slab thickness $\tau=5$ voxels.
Voxels are sorted lexicographically by $(s_n, c_n^y, c_n^x)$.
Within each slab, this ordering visits voxels in row-major order, providing locally consistent horizontal context while advancing sequentially from ground to canopy.
The reverse scan (top-down) used in the dual-path decoder is obtained by negating all keys.

\section{Hyperparameter Summary}
\label{sec:hyperparams}
Table~\ref{tab:hyperparams} collects all major hyperparameters of ForestMamba, grouped by component.
All values are shared across all seven forest regions unless stated otherwise.

\vspace{4pt}
\noindent\textbf{Encoder.}
The voxel size $v{=}0.2$\,m balances spatial resolution and memory: finer voxels recover more geometric detail but increase $N_v$ quadratically in the horizontal plane.
The slab thickness $\tau{=}5$ voxels was chosen so that each slab spans approximately 1\,m in height, which is finer than the typical inter-whorl spacing of coniferous trees but coarse enough to group ground and low-trunk voxels into a single contextual unit.
The Mamba state dimension $d_\text{state}{=}16$ and expansion factor $\text{expand}{=}1$ keep the encoder SSM lightweight; a larger state produced no measurable gain in our ablations.

\vspace{4pt}
\noindent\textbf{CHM.}
The two CHM resolutions, $r_f{=}0.3$\,m and $r_c{=}0.7$\,m, target the fine and coarse crown structures, respectively, covering both small suppressed trees and large dominant crowns.
The minimum peak height $h_\text{min}{=}1.5$\,m discards ground-level noise and shrubs that are not of interest as individual tree instances.
The minimum separation $d_\text{min}{=}1.0$\,m is set conservatively to allow closely spaced small trees to generate distinct seeds.
The cylinder radius $r_p{=}1.5$\,m for feature pooling was chosen to approximately match the median crown radius in the training set, so that each CHM query integrates features from the full crown footprint rather than just the treetop voxel.
The allometric parameters $(\alpha,\beta){=}(0.25, 0.5)$ were empirically tuned on the training split.
The exponent $\beta{=}0.5$ reflects the well-established ecological observation that crown radius grows as a sublinear power law of tree height~\cite{jucker2017allometric,pretzsch2009allometry}.
The scale coefficient $\alpha{=}0.25$ was selected to match the observed crown-to-height ratio in the FOR-instanceV2~\cite{xiang2025forestformer3d} training data.

\vspace{4pt}
\noindent\textbf{Decoder.}
The query budget $K{=}300$ was set to exceed the maximum number of tree instances in any single cylindrical patch (observed maximum ${\approx}230$), ensuring sufficient capacity without introducing excessive negative queries.
The number of decoder layers $L{=}6$ and model dimension $D'{=}256$ follow common practice in query-based segmentation~\cite{kolodiazhnyi2024oneformer3d,xiang2025forestformer3d}.
The kNN neighborhood size $\kappa{=}16$ provides a local aggregation window that covers roughly one crown diameter at the 0.2\,m voxel resolution, balancing receptive field size against computational cost.

\vspace{4pt}
\noindent\textbf{Training.}
We train for 3{,}000 epochs using AdamW~\cite{loshchilov2019decoupled} with a base learning rate of $10^{-4}$ and polynomial decay (power $0.9$), which we found more stable than cosine annealing for this dataset size.
Weight decay of $0.05$ and gradient clipping at a norm of $10$ prevent divergence during the early decoder training phase when query anchors are unstable.

\vspace{4pt}
\noindent\textbf{Loss weights.}
The semantic loss weight $\lambda_\text{sem}{=}0.2$ is kept small relative to the instance loss to avoid the encoder over-fitting to semantic boundaries at the expense of instance-discriminative features.
Within the instance loss, the Dice weight $w_\text{dice}{=}0.5$ is halved relative to BCE because Dice already provides a normalized overlap signal; using equal weights leads to over-penalization of small instances.
The discriminative margin $\delta_v{=}0.5$ and $\delta_d{=}1.5$ follow the original formulation~\cite{de2017semantic} and enforce a minimum inter-instance separation of $2\delta_d{=}3.0$ in embedding space.


\begin{table}[t]
\centering
\caption{Hyperparameter summary.}
\label{tab:hyperparams}
\setlength{\tabcolsep}{5pt}
\begin{tabular}{lll}
\toprule
Component & Parameter & Value \\
\midrule
\multirow{5}{*}{Encoder}
  & Voxel size $v$ & 0.2\,m \\
  & Slab thickness $\tau$ & 5 voxels \\
  & Mamba $d_{\text{state}}$ & 16 \\
  & Mamba expand & 1 \\
  & Mamba $d_{\text{conv}}$ & 4 \\
\midrule
\multirow{5}{*}{CHM}
  & Scales $\mathcal{R}$ & \{0.3\,m, 0.7\,m\} \\
  & Allometric $(\alpha,\beta)$ & (0.25, 0.5) \\
  & Min.\ peak height $h_{\min}$ & 1.5\,m \\
  & Min.\ separation $d_{\min}$ & 1.0\,m \\
  & Cylinder radius $r_p$ & 1.5\,m \\
\midrule
\multirow{6}{*}{Decoder}
  & Query budget $K$ & 300 \\
  & Decoder layers $L$ & 6 \\
  & Model dim.\ $D'$ & 256 \\
  & Mamba $d_{\text{state}}$ & 64 \\
  & FFN hidden dim. & 1024 \\
  & kNN neighbours $\kappa$ & 16 \\
\midrule
\multirow{6}{*}{Training}
  & Epochs & 3\,000 \\
  & Optimizer & AdamW~\cite{loshchilov2019decoupled} \\
  & Learning rate & $10^{-4}$ \\
  & Weight decay & 0.05 \\
  & LR decay & Polynomial (power 0.9) \\
  & Gradient clip & norm 10 \\
\midrule
\multirow{4}{*}{Loss weights}
  & $(\lambda_{\text{ins}},\lambda_{\text{sem}},\lambda_{\text{bin}},\lambda_{\text{dis}})$ & (1.0, 0.2, 1.0, 1.0) \\
  & $(w_{\text{cls}},w_{\text{bce}},w_{\text{dice}})$ & (1.0, 1.0, 0.5) \\
  & Disc.\ margins $(\delta_v,\delta_d)$ & (0.5, 1.5) \\
  & Matching IoU threshold $\theta$ & 0.5 \\
\bottomrule
\end{tabular}
\end{table}

\section{Failure Cases and Limitations}
\label{sec:failure}

\noindent\textbf{Tropical and multi-stemmed trees.}
In highly heterogeneous scenes such as YuChen (French Guiana), the CHM-based query initialization can fail for trees with irregular or multi-stemmed crowns that do not produce clear canopy peaks.
Additionally, dense understory vegetation may confuse the auxiliary branch, leading to extreme canopy overlap, as in parts of the BlueCat region; even the dual-path Mamba decoder can struggle to separate instances in such areas.

\vspace{5pt}
\noindent\textbf{Very dense canopies.}
In scenes with extreme canopy overlap, such as parts of the BlueCat region, even the dual-path Mamba decoder can struggle to separate instances with nearly identical spatial footprints.
In such cases, the NMS-based peak deduplication may merge distinct treetops, reducing recall.

\vspace{5pt}
\noindent\textbf{Encoder latency trade-off.}
As reported in Table~\ref{tab:complexity}, the Sparse Mamba encoder is moderately slower than the standard SpConvUNet backbone (246.5\,ms vs.\ 196.7\,ms) due to slab serialization and the Mamba scan.
This overhead is more than compensated by the $4.2\times$ decoder speedup, yielding an overall $3.0\times$ end-to-end gain.
Future work could adopt hardware-optimized Mamba kernels or fused CUDA implementations to further reduce encoder latency.

\vspace{5pt}
\noindent\textbf{Fixed allometric parameters.}
The suppression window parameters $(\alpha,\beta){=}(0.25, 0.5)$ are tuned on the training split and shared across all seven forest regions without per-region adaptation.
Although the power-law form is ecologically grounded~\cite{jucker2017allometric,pretzsch2009allometry}, a single pair of coefficients may not be optimal for all forest types. 
For instance, tropical multi-layered stands may require a smaller $\alpha$ to avoid over-suppression of subdominant crowns.
Learning region- or species-adaptive allometric parameters from data is a promising direction for future work.

\end{document}